# Exploring Modular Mobility: Industry Advancements, Research Trends, and Future Directions on Modular Autonomous Vehicles

Lanhang Ye, Toshiyuki Yamamoto

Institute of Materials and Systems for Sustainability, Nagoya University

ye.lanhang.n2@f.mail.nagoya-u.ac.jp

**Abstract**

Modular autonomous vehicles (MAVs) represent a transformative paradigm in the rapidly advancing field of autonomous vehicle technology. The integration of modularity offers numerous advantages, poised to reshape urban mobility systems and foster innovation in this emerging domain. Although publications on MAVs have only gained traction in the past five years, these pioneering efforts are critical for envisioning the future of modular mobility. This work provides a comprehensive review of industry and academic contributions to MAV development up to 2024, encompassing conceptualization, design, and applications in both passenger and logistics transport. The review systematically defines MAVs and outlines their technical framework, highlighting groundbreaking efforts in vehicular conceptualization, system design, and business models by the automotive industry and emerging mobility service providers. It also synthesizes academic research on key topics, including passenger and logistics transport, and their integration within future mobility ecosystems. The review concludes by identifying challenges, summarizing the current state of the art, and proposing future research directions to advance the development of modular autonomous mobility systems.

## 1. Introduction

The pursuit of efficient movement of people and goods has driven a new wave of technological innovation across the transportation sector. Over the past decade, transportation research has undergone a significant paradigm shift, with a growing focus on mobility technologies. This shift has been fueled by advancements in automotive technology, particularly the emergence of autonomous vehicles, which have garnered substantial interest from academia and industry. Technologies such as artificial intelligence and advanced automation have further enabled the development of diverse mobility solutions tailored to meet the needs of future societies.

Among these innovations, Modular Autonomous Vehicles (MAVs) have emerged as a particularly compelling concept, attracting attention from researchers and industry stakeholders. MAVs are envisioned as adaptable and flexible platforms that integrate autonomy with modularity, offering versatile solutions for a wide range of transportation demands. These vehicles hold the potential to address critical challenges such as urban congestion, environmental sustainability, and accessibility. By transforming transportation systems, MAVs promise to enhance mobility for both passengers and logistics.

This study undertakes a comprehensive systematic review of existing research on MAVs to provide a deeper understanding of modular mobility. While MAV research is still in its infancy, with most publications emerging in the past five years, it represents a promising direction for the future. The potential of this technology to revolutionize mobility systems makes it an essential focus for both academic and industrial exploration.

Our review synthesizes insights from diverse academic and industrial efforts, spanning the journey from conceptualization to real-world demonstrations. By analyzing contributions from various entities, we aim to illuminate the capabilities, challenges, and implications associated with MAVs. Furthermore, this investigation delves into the diverse applications of MAVs as explored in prior research and proposes future directions for their deployment. Ultimately, this work aspires to contribute to the vision of a futuristic transportation system that seamlessly integrates modular mobility technologies, advancing the realization of efficient, sustainable, and inclusive mobility solutions.

This paper is structured as follows: Section 2 introduces the concept of MAVs, detailing their technical framework with a focus on vehicle design and modularity types. Section 3 examines the early stages of modular mobility, highlighting contributions from automotive companies, startups, and design agencies. Section 4 reviews academic research efforts, categorizing them into key topics, including passenger transport systems, logistics, delivery services, co-modal transport systems, and pioneering studies in related areas. Finally, the review concludes by proposing future research directions and envisioning the evolution of modular mobility systems.

## 2. Concept and Classification

MAV integrate the concept of modularity into the ongoing evolution of autonomous vehicle technology. This modularity manifests in two primary dimensions: vehicular design and operational configuration. Vehicular design modularity involves designing vehicles with interchangeable modules tailored for specific purposes. A typical approach separates the self-driving unit (drive unit) from the functional capsule (body module). The drive unit serves as a universal component that can be paired with different body modules, enabling diverse applications such as passenger transport, cargo delivery, or specialized services. This design allows MAVs to adapt to various needs flexibly, optimizing resource utilization and improving overall efficiency. On the other hand, operational modularity focuses on coupling and decoupling individual modules dynamically based on operational requirements. For instance, modules can be configured to accommodate passengers, carry cargo, or deliver specialized services, with the ability to reconfigure systems as demands evolve. By integrating these two dimensions of modularity, MAVs offer notable advantages such as flexibility, scalability, efficiency, and customization.

### 2.1 Modularity Integration

#### 2.1.1. Separation of Driving Unit and Body Module

The driving unit and body module operate as distinct components, enabling flexibility in swapping body modules for different applications. For example, a single drive unit can alternately function as a passenger vehicle, a logistics carrier, or a specialized utility vehicle by coupling with appropriate modules.

#### 2.1.2 Integration of Driving Unit and Body Module

In this design, the driving unit and body module are unified into a single, cohesive structure. While it sacrifices the flexibility of swapping modules, this configuration can be more streamlined and efficient for use cases where modular separation is unnecessary or impractical.

## 2.2. Modularity Configuration

### 2.2.1. Physically Docked Operation

Modules are physically connected during operation, ensuring secure integration for power and data sharing. This configuration is further divided into:

a. **Shared Internal Space**: Enables free movement of passengers or cargo between connected modules.

b. **Separate Internal Space**: Maintains independent internal spaces for each module, limiting interaction between them.

### 2.2.2. Virtually Attached Operation

Modules maintain a small physical gap but operate in a coordinated manner. This approach enhances flexibility and adaptability, allowing modules to move independently while collaborating functionally.

## 3. Docking Methods

Docking systems in MAVs can be further categorized based on where and how the docking occurs:

a. **En-Route Docking**: Modules integrate seamlessly while in motion, allowing dynamic reconfiguration without halting operations.
b. **Station-Wise Docking:** Modules dock at designated stations where vehicles must remain stationary during the process.

The modularity configurations outlined above are grounded in pioneering research studies across various sectors, exploring operational features tailored for this emerging mobility technology. These studies, which span academia and industry, aim to design and envision the future of modular mobility systems. Tables 1 and 2 provide a detailed summary of the classifications, along with notable industry prototypes that exemplify these configurations.

Table 1 Technical categories of modularity in MAV

| | Modularity Configuration | Vehicular Operation | Unique operations | Industry prototype |
|---|---|---|---|---|
| Modular autonomous vehicle (MAV) | Separation of Driving Unit and Body Module | Physically Docked | Integration of driving unit and body module | U-Shift |
| | Integration of Driving Unit and Body Module | | Docking and undocking of Modules | Next Modular Mobility |
| | | Virtually Attached | Synchronizing movements between coupled modules | Ohmio Automotion |

Table 2 Industry prototypes of different types of MAV

| U-Shift | Next Modular Mobility | Ohmio Automotion |
|---|---|---|
| 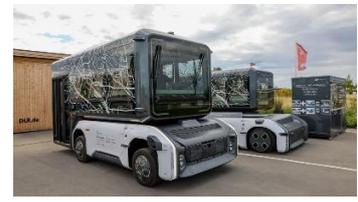 | 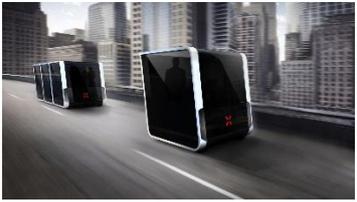 | 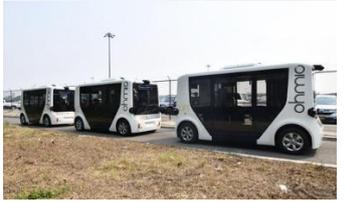 |

Photo credit: U-Shift, Next Modular Mobility, Ohmio Automotion

## 3. Industry pioneers

The concept of MAV has been pioneered by leading automobile manufacturers, innovative startups, and design agencies. Companies such as Mercedes-Benz, startups like Next Modular Mobility, and research organizations like the German Aerospace Center have laid the foundation for this transformative approach to transportation. Design agencies such as Mormedi have also contributed significantly, pushing the boundaries of vehicle conceptualization and modular design. Over the years, numerous prototype vehicles and demonstrations, including those by Next Modular Mobility, Ohmio and others, have showcased the potential applications of MAVs in both passenger and logistics transport.

In 2018, Mercedes-Benz introduced the groundbreaking modular concept vehicle, Vision Urbanetic, which revolutionized vehicle design by combining a self-driving, electrically powered chassis with interchangeable body modules for passenger and cargo transport. The innovative design allows modules to be swapped automatically or manually within minutes, significantly improving flexibility and operational efficiency. This approach, which separates the driving unit from the body module, was a landmark innovation that introduced modularity into autonomous vehicles, setting the stage for future MAV development.

The startup Next Modular Mobility has also been at the forefront of innovation. In 2019, Gecchelin and Webb (2019) presented a modular electric bus prototype featuring docking (coupling) and undocking (decoupling) in the longitudinal direction of multiple modules. These buses demonstrated seamless en-route docking, enabling modules to function as a cohesive unit. Each module is equipped with front and rear doors, facilitating passenger redistribution based on destination, which enhances operational efficiency. Building on this, Gecchelin and Spera (2022) patented the Next Modular Patented Self-Driving Vehicle (Pod), further advancing this modular mobility concept.

In 2020, the German Aerospace Center unveiled the U-Shift, a prototype designed to meet future urban mobility needs for both passenger and logistics applications. The U-Shift comprises two key components: capsule-shaped modules for transporting people or goods and a U-shaped autonomous drive unit. Potential use cases include on-demand shuttles, on-call buses, and round-the-clock distribution centers for goods. This innovative design, which combines modularity, electric propulsion, and autonomy, highlights the versatility of MAVs and their potential to transform urban mobility systems (Ulrich et al., 2019).

The design agency Mormedi introduced the 1O1 - Modular Mobility concept in 2022, an ambitious vision of urban mobility that integrates modularity to address diverse transportation needs. This design emphasizes versatility, allowing seamless transitions between roles such as passenger transport, on-demand services, and last-mile logistics. The 1O1 - Modular Mobility system reflects Mormedi's forward-thinking approach to creating efficient and integrated urban transportation solutions, shaping the future of mobility by combining adaptability and technological innovation.

In 2023, the New Zealand-based autonomous mobility company Ohmio Automotion showcased its three-vehicle platooning electric driverless shuttles. Each shuttle accommodates up to eight passengers, operates at a maximum speed of 23 mph, and maintains a 2-meter separation distance between vehicles. This demonstration exemplifies the integration of modularity into autonomous vehicle technology, offering a glimpse into the potential of MAVs for future passenger and freight mobility systems.

Over the years, a range of industrial pioneers, including automotive manufacturers, startups, and design agencies, have made significant strides in the development of modular mobility solutions. These efforts span vehicle conceptualization, system design, and real-world demonstrations, propelling the advancement of MAV technology. The evolution of autonomous vehicle systems has accelerated the progress of modular mobility, while academic research has focused on optimizing their applications and operations. Together, these initiatives are shaping a new era of efficient, adaptable, and sustainable transportation systems.

## 4. Pioneering Research efforts

The potential applications of MAV encompass a diverse range of transportation needs, including passenger transit and logistics services. This section categorizes prior research efforts into three primary domains: public transit for passenger transport, cargo and logistics delivery, and other related pioneering studies. Existing studies are reviewed within these categories and presented in chronological order based on their publication date and online availability.

### 4.1. Public transit for passenger transport

MAV technology holds significant promise in passenger transport due to its adaptability and flexibility. The ability of MAVs to couple and decouple multiple modules, either during transit or at designated stops, offers innovative solutions for addressing spatial and temporal demand imbalances in public transit systems. These modules can either physically dock for seamless integration or operate in close proximity while functioning independently, enhancing operational efficiency and service customization.

Research in this domain has primarily focused on designing modular transit systems, optimizing operational strategies, and exploring practical applications of MAVs. Key areas of investigation include system design, scheduling, routing, vehicle formation, and the integration of advanced management strategies. This section reviews existing studies on MAV applications in passenger transport, highlighting three core aspects:

    a. **System Design**: Early studies introduced the fundamental features of modular transit systems,

such as adjustable capacity, join-and-detach capabilities, and en-route transfer mechanisms. These features enable transit systems to adapt dynamically to fluctuating passenger demands.
   b. **Operational Strategies**: Subsequent research emphasized the optimization of operational processes, including scheduling, routing, and charging operations for modular vehicles. These studies aimed to enhance the efficiency and effectiveness of public transit systems by leveraging the unique capabilities of MAVs.
   c. **Field Applications**: MAVs have also been proposed for specialized passenger transport scenarios, such as emergency medical services, metro systems, and airport shuttles. These applications highlight the versatility of MAV technology in addressing diverse transportation needs.

The existing literature on modular vehicles in passenger transport can be categorized into six main areas based on the key features and focus of the studies. Early research introduced groundbreaking attributes of modular transit, such as adjustable capacity, join-and-detach functionality, and en-route transfer. These foundational features have set the stage for subsequent studies, which have focused on designing advanced transit systems that leverage these capabilities while optimizing operational processes. Key areas of optimization include scheduling, vehicle formation, routing, and charging strategies.

Additionally, some studies have extended the application of modular vehicles to specialized contexts, including emergency medical services, metro systems, and airport shuttles. These explorations underscore the versatility and potential of MAV technology beyond conventional transit scenarios. It is important to note that the features defined in different categories are often interrelated. For example, the join-and-detach capability enables en-route transfer, but its presence does not inherently imply that en-route transfer will be implemented. Such nuanced distinctions highlight the flexibility of MAV systems in addressing varied operational needs.

Table 3 provides a detailed summary of recent literature across these categories, outlining their primary focus and key features, offering insights into the evolution and application of modular transit systems.

Table 3 Literature on application of MAV in passenger transport system

| Research Category | Literature | Key feature |
|---|---|---|
| Adjustable capacity in dispatching | Chen et. al (2019, 2020) | Shuttle system of oversaturate traffic |
| | Shi et al. (2020) | Fluctuating demand in corridor transit |
| | Dakic et al. (2021) | Flexible dispatching system |
| | Shi and Li (2021) | |
| | Romea and Estrada (2021) | Analytical analysis |
| | Huang et al. (2025) | Hybrid transit system |
| Join-and-detach capability | Chen and Li (2021) | Station-wise docking |
| | Chen et al. (2022) | En-route coupling and decoupling |
| | Khan et al. (2023) | |
| | Tang et al. (2024) | |
| | Khan and Menéndez (2023) | En-route splitting |
| En-route passenger transfer | Caros and Chow (2021) | En-route transfer |
| | Gong et al. (2021) | |
| | Fu and Chow (2022, 2023) | |
| | Zou et al. (2024) | |
| | Lin et al. (2024) | |
| | Cheng et al. (2024) | |
| | Wu et al. (2021) | In-motion transfer |
| | Liu et al. (2024) | |
| Modular transit system innovations | Zhang et al. (2020) | First- and last-mile connections |
| | Zermasli et al. (2023) | |
| | Yi et al. (2024) | |
| | Luo et al. (2024) | |
| | Pei et al. (2021) | Modular transit network system |
| | Liu et al. (2021) | Flex-route transit services |
| | Tian et al. (2022) | Modular transit system planning |
| | Khan and Menéndez (2022) | Stop-Less modular bus service |
| | Fu (2023) | Modular and electric microtransit |
| | Guo et al. (2023a) | Modular customized bus system |
| | Khan and Menéndez (2024) | Network-wide seamless bus service |
| Joint optimization of operation | Liu et.al (2020) | Minimum fleet size |
| | Ji et al. (2021) | Scheduling |
| | Tian et al. (2023, 2025) | Scheduling and vehicle formation |
| | Guo et al. (2023b) | Routing, charging, and assignment |
| | Wang et al. (2023) | Routing and timetabling |
| | Gao et al. (2023) | Scheduling, formation, charging |
| | Liu et al. (2023) | Timetable, formation, and scheduling |
| | Xia et al. (2023, 2024) | Timetable and vehicle formation |
| | Zhang et al. (2024) | Skip-stop strategy |
| | Yi et al.(2024) | Scheduling and charging |
| | Chang et al. (2024) | Dispatching and charging |
| Application in diverse fields | Hannoun and Menendez (2022) | Emergency medical services |
| | Pei et al. (2023) | Metro system |
| | Oargă et al. (2024a) | Airport shuttle service |
| | Oargă et al. (2024b) | |
| | Wang et al. (2024) | Multimodal transit system |

The following sections provide a detailed overview of each category, presented in chronological order based on the online publication dates of the corresponding studies.

### 4.1.1. Adjustable capacity in dispatching

Early research into MAV focused on utilizing their adjustable capacity to address supply-demand imbalances, particularly in corridor-based public transport settings. These studies sought to accommodate fluctuating demand during peak and off-peak hours by dispatching modular buses with varying capacities, achieved through the coupling and decoupling of bus modules.

Chen et al. (2019, 2020) introduced discrete and continuous modeling methods to design shuttle operations using MAVs in oversaturated traffic conditions. Their work represents a pioneering effort to integrate MAVs into urban mass transit systems, providing foundational insights into their operational dynamics.

Shi et al. (2020) developed a linear optimization model for modular transit operations, emphasizing cost efficiency. Their findings demonstrated the potential of MAVs to minimize overall system costs by leveraging adjustable vehicle capacities to better align supply with demand.

Dakic et al. (2021) proposed a flexible dispatching system for managing a fleet of fully automated modular bus units alongside conventional buses. This system dynamically adjusts the number of modular units and their dispatch frequencies to meet passenger demand, reduce operating costs, and enhance mobility. Their results showed significant improvements in total system cost by optimizing modular bus configurations and scheduling.

Shi and Li (2021) explored the operations of a multistep transit corridor where vehicle capacities could dynamically vary to address fluctuating passenger demands across space and time. By formulating a mixed-integer linear programming model, they optimized headways and vehicle capacities, achieving cost-efficient solutions for transit corridors.

Romea and Estrada (2021) investigated the potential of an Autonomous Driving Modular Bus capable of adjusting transport capacity through pod coupling and decoupling. Their analysis revealed that this technology could substantially reduce user travel times and operating costs, particularly on high-demand routes, highlighting its viability as a cost-effective solution for future urban transit systems.

Huang et al. (2025) developed a bi-level programming model to optimize hybrid transit systems with modular autonomous and conventional buses, showing that integrating modular buses of varying capacities enhances sustainability by reducing operating costs (25.1%), environmental costs (27.72%), passenger waiting times, and carbon emissions.

### 4.1.2. Join-and-detach capability

Research on modular vehicles has increasingly focused on their ability to join or detach modules, either at stations or en route. This capability addresses challenges such as bus bunching by enabling modules to decouple during operation through station-wise docking or en-route splitting. Compared to traditional transit systems, this innovative feature introduces greater flexibility and helps mitigate common issues like inconsistent headways and service reliability in public transportation.

Chen and Li (2021) explored the station-wise docking and undocking capabilities of MAVs in corridor-based transit systems. Their study highlighted the benefits of shifting from fixed-capacity designs to MAV-based configurations, demonstrating reductions in system costs and improvements in operational efficiency. This research laid the groundwork for adopting MAVs in urban mass transit systems.

Building on this foundation, Chen et al. (2022) proposed a continuous modeling approach to design MAV-based transit corridors. Their model provided near-optimal solutions for operational strategies, ensuring efficient performance and practical applicability of MAV-based systems.

Khan et al. (2023) investigated the application of MAV technology to address the persistent issue of bus bunching. Their study leveraged the en-route coupling and decoupling capabilities of MAV buses as a novel alternative to conventional strategies like bus-holding, stop-skipping, and insertion-based methods. The results demonstrated that this approach significantly reduces travel costs, outperforming benchmarks for high-demand bus lines by over twofold.

Tang et al. (2024) introduced an innovative hybrid public transit service that utilizes coupled MAV fleets for regular routes while accommodating door-to-door requests with decoupled MAV units in a deviated service area. This approach showcases the potential for MAV-enabled flexible transit services to enhance accessibility and adaptability compared to traditional fixed-route systems.

Khan and Menéndez (2023) proposed a hybrid splitting-holding strategy for the operation of autonomous modular buses to mitigate bus bunching. Their study demonstrated that this strategy outperforms standalone splitting and skipping-based approaches, significantly reducing both the average travel costs and their variability. The results underscore the potential of MAVs with in-motion transfer capabilities to optimize public transit operations.

### 4.1.3. En-route passenger transfer

Building on the join-and-detach capability of MAV modules, researchers have further explored en-route or in-motion passenger transfers, proposing innovative service models to revolutionize transit systems. These advancements enhance flexibility, service quality, and efficiency in public transportation, paving the way for cutting-edge MAV transit systems with futuristic features.

Caros and Chow (2021) evaluated the performance of MAV-based mobility services capable of en-route passenger transfers. They examined three operating strategies: door-to-door service, first/last mile service, and hub-and-spoke service. Their study revealed that en-route transfer capabilities could enhance service quality, alleviate congestion, and address the last-mile challenge effectively.

Gong et al. (2021) designed a transfer-based customized bus network using a modular bus fleet, focusing on optimizing passenger-route assignments. Their findings demonstrated significant efficiency gains and highlighted the potential of modular buses in enhancing customized transit services.

Wu et al. (2021) introduced a modular, adaptive, and autonomous transit system that adapts to dynamic traffic demands through in-motion transfer operations, bypassing the limitations of fixed routes and timetables. Their work envisioned a futuristic public transit system that maximizes flexibility and responsiveness.

Fu and Chow (2022) developed a mixed-integer linear programming (MILP) model for microtransit systems leveraging modular autonomous vehicles with synchronized en-route transfers. Their findings indicated that such systems could reduce total costs by an average of 10%, with savings reaching up to 19.6%, underscoring the cost-effectiveness of modular MAV solutions.

Fu and Chow (2023) extended their research to demand-responsive MAV fleets capable of en-route passenger transfers before splitting. Their study demonstrated potential savings of up to 52% in vehicle travel costs, 41% in passenger service time, and 29% in total costs compared to existing on-demand mobility services.

Zou et al. (2024) proposed a Modular Electrified Transit system for corridor areas, integrating mainline and feeder transit connected by en-route transfers through docking sections. The system demonstrated a 24.4% reduction in electricity consumption while improving operational efficiency and flexibility.

Lin et al. (2024) explored the use of modular vehicles to address bus bunching and enhance system flexibility through adaptable boarding capacities. Their findings highlighted the potential for multiple passenger queues to serve each module simultaneously, improving boarding efficiency and system performance.

Cheng et al. (2024) examined an Autonomous Modular Public Transit system operating on a grid network. Featuring pod joining, disjoining, and en-route passenger transfers, the system demonstrated significant cost reductions, especially in low-demand scenarios. Compared to traditional fixed-route transit services, this innovative approach showcased superior demand responsiveness and operational efficiency.

Liu et al. (2024) introduced a novel strategy for alleviating bus bunching in transit systems using MAVs. By integrating continuum approximation and stochastic optimization with a customized deep Q-network algorithm, the study optimized MAV splitting and merging operations. The results highlighted superior performance in mitigating bus bunching compared to conventional strategies across various scenarios.

### 4.1.4. Modular transit system innovations

Building on the capabilities of MAVs, several pioneering efforts have explored innovative transit systems that leverage their modular and autonomous features. These cutting-edge systems aim to address key challenges in public transportation, including first- and last-mile connectivity, flexible transit routing, demand-responsive operations, and streamlined, stop-less service.

Zhang et al. (2020) presented a mathematical model for a modular transit system operating on a time-expanded network. They analyzed various vehicle configurations and demonstrated that modular buses can be deployed not only for local trips but also for efficient long-distance travel. By providing convenient transfers to main modules covering the bulk of longer journeys, this system improved overall vehicle utilization and served a higher number of passengers compared to traditional door-to-door services.

Pei et al. (2021) introduced the Modular Transit Network System (MTNS) concept, utilizing MAV technology to dynamically adjust vehicle capacity through the docking and undocking of modular pods. To optimize the system, they developed a mixed-integer nonlinear programming model that balances vehicle operation costs with passenger trip time costs. Their findings indicate that the MTNS can substantially reduce system-wide costs by offering flexible capacity aligned with spatial demand variations.

Liu et al. (2021) proposed a flex-route service design that integrates MAVs and self-adaptive capacity with service flexibility. Their model responds to time- and space-dependent demand, enhancing both operational efficiency and passenger experience. Numerical results from their study suggest that these strategies can effectively reduce both vehicle operation and passenger travel costs.

Focusing on strategic planning, Tian et al. (2022) developed an optimization model to plan modular vehicle transit systems that minimizes costs for operators and passengers while considering station placement, capacity, and vehicle formation. Their approach was validated through a case study on Singapore's proposed dynamic autonomous road transit (DART) system.

Khan and Menéndez (2022) introduced a service paradigm for non-stop modular buses, where supplementary units detach and reattach at stops to accommodate boarding and alighting passengers without slowing the main vehicle. Their case study demonstrated that such an approach can reduce average travel costs by about 15–20% compared to conventional operations.

In a more comprehensive study, Fu (2023) presented a doctoral dissertation on the operational design and dynamic charging strategies for modular, electric microtransit systems. This work outlines systematic planning methods and operational frameworks for real-world implementation, offering valuable insights into best practices for deploying MAV-based transit services.

Zermasli et al. (2023) focused on integrating modular autonomous buses into feeder service operations. By enabling on-the-fly assembly and disassembly of pods for en-route transfers, their design facilitated efficient feeder connections to high-capacity modes such as metro systems, delivering cost savings and improved service quality in low-demand areas.

Guo et al. (2023) studied scheduling and dispatching strategies for on-demand, MAV-based bus services. Their findings highlight the system's potential to achieve high resource utilization and environmental benefits, improving overall service efficiency.

Yi et al. (2024) investigated an electric public transit system employing MAVs capable of seamless mid-route coupling and decoupling. Their work addressed the joint optimization of energy consumption and passenger travel costs, demonstrating that such a system can outperform conventional public transit and taxi services in multiple performance dimensions.

Khan and Menéndez (2024) further developed a network-level "seamless" bus concept, in which modular buses interact with different lines by exchanging pods. Through simulations, they showed that a citywide deployment of this concept could reduce travel costs by as much as 35% relative to conventional bus services.

Lastly, Luo et al. (2024) proposed a Modular Feeder Transit system featuring decoupled MAVs to tackle the first/last-mile problem. Their optimal design framework revealed over 10% cost savings versus traditional fixed-capacity feeder bus services, while substantially enhancing both operational flexibility and overall performance.

### 4.1.5. Joint optimization of operations

Studies in this category focus on the joint optimization of operational aspects—such as routing, scheduling, vehicle formation, dispatching, and charging—in modular transit systems. By leveraging MAVs' adaptive capacity and their ability to dynamically join and detach modules, researchers aim to overcome the limitations of traditional public transit systems, including congestion, rigid operations, and underutilized resources. Through integrated optimization models, these works seek to identify strategies that simultaneously address operator and passenger costs, increase capacity utilization, improve reliability, and reduce waiting times. The overarching goal is to design holistic, adaptive, and cost-effective MAV-based transit services that balance real-time demand fluctuations, energy constraints, and the unique structural flexibility afforded by modular technologies.

Liu et al. (2020) developed an extended deficit function model to estimate the minimum number of modules needed to cover scheduled services in an autonomous modular public transit network. Their research laid the groundwork for large-scale, multi-line, and multi-terminal systems, demonstrating how well-structured modular operations can reduce fleet requirements.

Ji et al. (2021) examined modular autonomous vehicle scheduling strategies for meeting time-varying passenger demand on transit routes. They found that MAV-based operations can significantly increase capacity utilization and decrease total passenger waiting times by 12.62%, underscoring the advantage of modular adaptability.

Tian et al. (2023, 2025) investigated optimal scheduling and vehicle formation at specialized stations for module assembly and disassembly. Their research highlighted the effectiveness of modular vehicles in adapting to fluctuating demand and minimizing operator and passenger costs. They developed both exact and heuristic solution methods to support real-time operations of modular-vehicle transit services.

Guo et al. (2023) integrated travel demand prediction into a modular, customized bus system powered by MAVs to proactively optimize operations. Results highlight the importance of module capacity and time-sensitive reconfiguration, showing robust operational cost efficiency and improved responsiveness to evolving travel patterns.

Wang et al. (2023) investigated timetable optimization for modular public transport, incorporating coupling and decoupling capabilities at stops. Their findings suggest that modularity enhances service reliability and mitigates situations where passengers are left behind due to capacity constraints.

Gao et al. (2023) introduced battery capacity and charging considerations into modular bus scheduling. Their study indicated that flexible, battery-powered MAVs can achieve up to a 25% reduction in operating costs compared to conventional electric bus fleets, due to on-demand capacity adjustments.

Liu et al. (2023) jointly optimized timetables, vehicle formations, and scheduling operations for an MAV-based flexible transit system. Their research confirmed that such integrated optimization yields lower total system costs and improves vehicle utilization rates relative to traditional systems.

Xia et al. (2023, 2024) addressed the mismatch between supply and demand dynamics in bus operations using modular vehicles. By jointly designing timetables and dynamically allocating modular capacity, they reduced both passenger and operator costs compared to fixed-capacity operations, effectively managing uncertainty and demand fluctuations.

Zhang et al. (2024) coupled MAV formation strategies with a skip-stop operational design. Their approach cut total system costs by 9.87% to 32.09% and passenger travel costs by 17.92% to 38.54%, underscoring the synergy between flexible MAV operations and adaptive network design.

Yi et al. (2024) optimized charging and route schedules for modular bus systems by introducing module substitution and in-motion charging strategies, developing mathematical models to minimize total costs, and showcasing cost-efficient solutions and operational insights through real-world case studies.

Chang et al. (2024) integrated dispatching, charging plans, and charging infrastructure optimization for an autonomous modular bus system. Their findings show a 23.85% decrease in operating costs and a 5.92% reduction in energy consumption compared to standard bus routes, further demonstrating the holistic benefits of joint optimization in MAV-based systems.

### 4.1.6. Applications in diverse fields

The final category explores the application of modular vehicle technology across a variety of specialized domains. Beyond their integration into public transit systems, researchers have explored the use of MAVs in scenarios such as emergency medical services, metro networks, airport shuttles, and as innovative travel modes within multimodal transportation systems.

Hannoun and Menendez (2022) introduced a modular vehicle system tailored for emergency medical services. Their approach involved designing modular units capable of coupling and decoupling to facilitate patient transfer between modules during transport. By employing a mathematical programming model, they optimized assignment decisions in a deterministic setting. This study underscores the adaptability of MAV technology in critical healthcare scenarios, demonstrating its potential to enhance operational efficiency in emergency medical services.

Pei et al. (2023) extended the modular transit concept to metro networks. Their research introduced metro fleets designed for dynamic disassembly and assembly into identical modules at metro terminals. This innovative approach showcased the potential of MAV technology to yield economic, low-carbon, and ecological benefits, making it a viable option for sustainable urban rail transit systems.

Oargă et al. (2024a) explored the application of MAV systems in airport shuttle services. Simulations conducted within the context of airport operations revealed that MAVs, as a connecting mode, can significantly improve energy efficiency compared to conventional solutions. Their findings demonstrated the practicality and benefits of integrating MAVs into airport transit networks.

In a subsequent study, Oargă et al. (2024b) performed a comparative analysis of energy efficiency between battery electric buses and MAVs. Their research highlighted MAVs' ability to optimize passenger capacity, reduce travel times, and decrease energy consumption. By dynamically assigning routes based on passenger demand and travel distance, MAVs exhibited superior efficiency, showcasing their potential to revolutionize transit systems with sustainable operational strategies.

Wang et al. (2024) tackled the challenge of accommodating personalized travel needs within complex multimodal transportation systems. They formulated and solved a Heterogeneous Demand Traffic Assignment Problem (HD-TAP), incorporating modular Shared Autonomous Vehicles (SAVs) to address diverse traveler preferences. Their findings emphasized the importance of balancing SAV supply to sustain public transit usage while addressing individual mobility needs, positioning MAVs as an integral component of future multimodal transportation networks.

**4.2. Logistic, delivery, and co-modal transport system**

In cargo and logistics transportation, the concept of modularity has been widely explored, with the Automated Guided Vehicle (AGV) serving as a comparable example. AGVs utilize modular designs to transport various types of cargo across diverse scenarios, including industrial sites, warehouses, and ports. For instance, de Oliveira et al. (2019) introduced a modular autonomous vehicle design specifically tailored for carrying heavy loads within industrial parks. Their research highlighted the integration of modularity into vehicle design to address mobility service needs unique to cargo transport and handling. A prominent application of MAV technology is in container transport systems, where uniformly sized cargo modules are optimized to fit specific types of vehicles. These systems represent an advanced level of modularity, particularly within the domain of freight logistics. However, such applications fall outside the scope of this review. Instead, the focus here is on integrating modularity into autonomous vehicles intended for everyday and general-purpose logistics operations.

Table 4 provides a summary of recent efforts to implement MAV innovations across a range of freight transport applications. These include military logistics, last-mile package delivery, baggage and cargo transportation, and the development of co-modal frameworks capable of serving both passenger and logistics needs. These innovations demonstrate the versatility of MAVs and their potential to transform traditional freight systems into more adaptive, efficient, and integrated transport solutions.

Table 4 Literature on application of MAV in logistic and delivery service

| Research Category | Literature | Key feature |
| --- | --- | --- |
| Logistic and delivery services | Rezgui et al. (2019) | Last-mile package delivery |
| | Shafiee et al. (2024) | |
| | Li and Epureanu (2020) | Military logistics |
| | Shi et al. (2024) | Baggage transport for air transport |
| | Zhou et al. (2024) | Cargo delivery |
| Co-modal transport system | Lin et al. (2022) | Co-modal system design |
| | Hatzenbühler et al. (2023) | Routing problem |
| | Lin and Zhang (2024) | Scale operation |

### 4.2.1. Logistic and delivery services

The application of MAV technology in logistics and delivery shows immense potential for revolutionizing future logistics systems, particularly in last-mile delivery services. These services, which transport parcels from distribution centers to customers' doorsteps, are both labor-intensive and resource demanding. However, they remain a critical component of the logistics chain. MAVs, with their modular design, enable flexible configurations to accommodate varying parcel sizes and delivery routes, improving operational efficiency and reducing delivery times.

Rezgui et al. (2019) proposed an optimal routing strategy for modular electric vehicles in last-mile urban freight delivery, addressing battery limitations through modular capacity adjustments and utilizing a Variable Neighborhood Search (VNS) approach to minimize costs, outperforming existing methods in experimental benchmarks.

Li and Epureanu (2020) conducted an agent-based simulation study to evaluate the performance of modular autonomous vehicle fleets in hostile military scenarios. They highlighted several strategic advantages of MAVs, such as adaptability, learning-based decision-making, and reconfigurability, which outperformed conventional fleets in terms of win rate, damage recovery, and unpredictability. This study underscores the potential of modular vehicles beyond traditional logistics, emphasizing their transformative possibilities across diverse applications.

Shafiee et al. (2024) assessed the effectiveness and operational costs of MAV systems in last-mile package delivery. Their study revealed that while MAV train operations slightly increased waiting times, they significantly reduced overall operational costs compared to traditional depot-based methods using trucks. This cost reduction was particularly evident in areas with high demand variability. The findings highlight MAV technology's potential to lower transportation costs while maintaining delivery times and service levels.

Shi et al. (2024) evaluated the feasibility of using MAVs for baggage transportation for departing flights. Their study indicated that this innovative technology could effectively reduce greenhouse gas emissions when properly managed, as MAVs allow for flexible capacity adjustments by assembling or disassembling detachable units.

Zhou et al. (2024) explored modular vehicle applications in cargo delivery, where vehicles can dock and split en route. Using a vehicle routing model that accounted for the unique docking and splitting capabilities of MAVs, they found that modular vehicles could reduce delivery costs by approximately 5% compared to traditional methods. This study demonstrates the significant potential of MAV technology to optimize delivery operations and reduce overall costs in cargo logistics.

### 4.2.2. Integrated passenger and freight transport

The modularity enabled by MAV technology provides a versatile framework for addressing the evolving demands of both passenger and freight mobility. Its scalability allows for flexible deployment, including the ability to adjust fleet size and vehicle configurations to accommodate changing transportation needs. A particularly promising application of MAVs lies in integrating passenger and freight transport systems, enabling a holistic and efficient mobility solution for future urban environments. Several pioneering studies have explored this concept, demonstrating the potential benefits of a co-modal transportation system.

Lin et al. (2022) introduced an Autonomous Modular Vehicle Technology (AMVT) bimodality system that utilized MAVs for integrated public transit and last-mile logistics services. Their study highlighted research challenges, opportunities, and approaches associated with adopting such a system in real-world scenarios. They concluded that AMVT has significant potential to reshape urban mobility but emphasized the importance of understanding the trade-offs inherent in modular and co-modal designs for mobility service systems.

Hatzenbühler et al. (2023) investigated the routing problem for a combined passenger and freight transport system using MAVs. By integrating passenger and freight demand, their study demonstrated the system's potential to reduce total costs by over 50%, decrease empty vehicle kilometers by more than 60%, and shorten trip durations by over 60%. These findings underscore the efficiency and sustainability advantages of integrated modular mobility systems.

Lin and Zhang (2024) proposed a modular transit system designed to accommodate both passenger and freight transport. This co-modal system dynamically adjusts vehicle capacity by docking and undocking modules and reallocating space between passengers and freight at intermediate stations to meet fluctuating demands. They formulated an optimization problem to manage these adjustments and developed a two-stage algorithm to provide near-optimal solutions for large-scale scenarios within an acceptable timeframe. Their work highlights the practicality and operational efficiency of co-modal MAV systems in real-world applications.

These studies collectively reveal the transformative potential of integrating passenger and freight transport through MAV technology, offering a path toward more sustainable, adaptable, and efficient urban transportation systems.

### 4.3. Pioneering studies in related fields

The distinctive operational processes of MAVs, including coupling and decoupling—commonly referred to as docking and splitting—offer unprecedented flexibility in vehicle configuration and operation. These processes allow modules to connect or disconnect as needed, enabling the creation of unified MAV trains or independent operation of individual modules. Docking typically involves securely connecting modules to function as a single unit, while splitting allows modules to separate and operate autonomously. These operations can occur either en route or at designated stops, depending on the specific technical configurations and requirements of the MAVs. This adaptability facilitates real-time adjustments to the vehicle's configuration, improving responsiveness to varying demands and conditions.

The innovative capabilities of docking and splitting not only enhance the operational flexibility of MAVs but also open up new research opportunities, particularly in traffic flow modeling and simulation. Researchers can investigate the impact of these processes on traffic dynamics, vehicle interactions, and overall system performance, providing valuable insights into the future of modular mobility systems. Table 5 presents several pioneer studies in the related field:

Table 5 Literature of other pioneering efforts on modular mobility

| Research Category | Literature | Key feature |
|---|---|---|
| Trajectory planning | Li and Li (2022, 2023); Han et al. (2024) | Docking and split operation |
| Traffic flow modeling | Ye and Yamamoto (2024) | Mixed traffic flow dynamics |
| Public perceptions and acceptance | Rejali et al. (2024) | Public acceptance |
| System equilibrium | Li et al. (2025) | Mixed traffic system |

Li and Li (2022, 2023) introduced a trajectory planning and optimization framework for MAVs during docking and splitting operations. Their solution focused on improving ride comfort and fuel efficiency while ensuring efficient transitions between long MAV trains and shorter modular configurations.

Han et al. (2024) developed a hierarchical docking planning model using Nonlinear Model Predictive Control to address the challenges associated with the physical docking of modular buses. Their model incorporated objective functions, docking constraints, and collision avoidance measures to optimize performance across various docking scenarios.

Ye and Yamamoto (2024) proposed a simulation framework for mixed traffic flow systems, integrating MAV docking operations and their collective functionality as arbitrarily formed trains of varying modular sizes. Their study analyzed the effects of different traffic demand levels and MAV penetration rates on traffic flow dynamics, offering a detailed perspective on how MAV technology could reshape future traffic systems and influence traffic flow theory.

Rejali et al. (2024) conducted a survey on public perceptions and acceptance of autonomous modular transit. Their findings revealed that most respondents were receptive to MAV technology, largely due to its perceived usefulness and potential benefits.

Li et al. (2025) introduced a bi-mode bottleneck model to analyze the equilibrium of systems with autonomous modular buses (AMBs) and autonomous private vehicles (APVs), revealing their potential to reduce societal costs, alleviate congestion, and eventually dominate transportation as AMB technology matures.

By exploring both microscopic and macroscopic dimensions of MAV systems, researchers aim to optimize traffic management strategies, enhance road safety, and improve the efficiency of mixed traffic environments where MAVs and conventional vehicles coexist. These pioneering efforts establish a foundation for seamlessly integrating MAVs into the broader transportation ecosystem, driving innovative and sustainable traffic solutions.

## 5. Conclusions and future research directions

This work provides a comprehensive overview of advancements in MAV technology, documenting progress from industry and academia up to 2024. As a relatively new research domain, the concept of MAVs gained prominence following industry innovations by pioneers like Mercedes-Benz and Next

Modular Mobility. With most studies emerging in the past five years, we anticipate a substantial increase in research activity in the coming years. This review serves as an early milestone, capturing the pioneering efforts that lay the foundation for modular mobility and its transformative potential in next-generation automobile and mobility systems.

MAVs hold significant promise in shaping the future of Mobility-as-a-Service (MaaS) and Logistics-as-a-Service (LaaS) frameworks. By facilitating the integration of co-modal systems capable of serving both passenger and freight transportation needs, MAVs offer a sustainable pathway to reduce environmental impacts, optimize transportation networks, and promote resource sharing. These developments could redefine urban mobility systems and logistics networks while aligning with global sustainability goals.

This timely review highlights the initial efforts toward modular mobility, presenting state-of-the-art advancements, identifying future research directions, and fostering a deeper understanding of this evolving field. While most existing research focuses on passenger transit systems and modular extensions of current public transit frameworks, modular mobility remains at an early developmental stage. Future advancements require a multidisciplinary approach to address the complexities inherent in planning, operations, control, economics, environmental sustainability, user experience, infrastructure design, and technological integration.

To advance modular mobility systems, the following areas warrant focused research efforts:

1. Integration of Passenger and Goods Transport: Develop frameworks and technologies for seamless co-modal systems that cater to both passenger and freight transport needs, enabling dynamic and efficient utilization of resources.

2. Scalability and Interoperability: Design scalable modular systems capable of adapting to various use cases and ensure interoperability across diverse transportation networks and platforms.

3. Technological and Infrastructural Adaptations: Explore necessary advancements in infrastructure and vehicle technology to support the adoption of MAV systems in practical, real-world contexts.

4. Optimization of Energy Efficiency: Investigate energy optimization strategies, including renewable energy integration, battery management, and operation scheduling to enhance the sustainability of MAV systems.

5. Economic and Environmental Impact Assessment: Conduct comprehensive evaluations of economic feasibility and environmental impacts to support the sustainable deployment of modular mobility solutions.

6. User Acceptance and Policy Frameworks: Examine user behavior, preferences, and acceptance of MAVs while developing supportive policies and regulatory frameworks to foster widespread adoption.

Before large-scale deployment, comprehensive evaluations of MAV technology are essential,

including assessments of technical readiness, application scenarios, and compatibility with existing systems. Successful implementation will require close collaboration across multiple sectors, involving stakeholders from urban planning, regulatory bodies, the automobile industry, engineering, and investment communities. These collaborations will address challenges and leverage opportunities to ensure the efficient deployment of modular mobility systems.

This work envisions a future paradigm where MAV technology seamlessly integrates passenger transportation and logistics networks. Such a unified, service-oriented, on-demand platform would provide flexible and efficient mobility options for both people and goods. This vision supports a shift away from private car ownership toward shared, sustainable transportation systems. The resulting benefits include cost reduction, enhanced efficiency of transportation networks, and improved societal well-being, paving the way for a smarter, more connected, and environmentally responsible future.

## Reference

Caros, N. S., & Chow, J. Y. (2021). Day-to-day market evaluation of modular autonomous vehicle fleet operations with en-route transfers. Transportmetrica B: Transport Dynamics, 9(1), 109-133.

Chang, A., Cong, Y., Wang, C., & Bie, Y. (2024). Optimal Vehicle Scheduling and Charging Infrastructure Planning for Autonomous Modular Transit System. Sustainability, 16(8), 3316.

Chen, Z., & Li, X. (2021). Designing corridor systems with modular autonomous vehicles enabling station-wise docking: Discrete modeling method. Transportation Research Part E: Logistics and Transportation Review, 152, 102388.

Chen, Z., Li, X., & Qu, X. (2022). A continuous model for designing corridor systems with modular autonomous vehicles enabling station-wise docking. Transportation Science, 56(1), 1-30.

Chen, Z., Li, X., & Zhou, X. (2019). Operational design for shuttle systems with modular vehicles under oversaturated traffic: Discrete modeling method. Transportation Research Part B: Methodological, 122, 1-19.

Chen, Z., Li, X., & Zhou, X. (2020). Operational design for shuttle systems with modular vehicles under oversaturated traffic: Continuous modeling method. Transportation Research Part B: Methodological, 132, 76-100.

Cheng, X., Nie, Y. M., & Lin, J. (2024). An Autonomous Modular Public Transit service. Transportation Research Part C: Emerging Technologies, 104746.

Dakic, I., Yang, K., Menendez, M., & Chow, J. Y. (2021). On the design of an optimal flexible bus dispatching system with modular bus units: Using the three-dimensional macroscopic fundamental diagram. Transportation Research Part B: Methodological, 148, 38-59.

de Oliveira, H. D. B. C. L., Campilho, R. D. S. G., & Silva, F. J. G. (2019). Design of a modular solution for an autonomous vehicle for cargo transport and handling. Procedia Manufacturing, 38, 991-999.

Fu, Z. (2023). Operation Design for Modular and Electric Microtransit (Doctoral dissertation, ProQuest Dissertations Publishing).

Fu, Z., & Chow, J. Y. (2022). The pickup and delivery problem with synchronized en-route transfers for microtransit planning. Transportation Research Part E: Logistics and Transportation Review, 157, 102562.

Fu, Z., & Chow, J. Y. (2023). Dial-a-ride problem with modular platooning and en-route transfers. Transportation Research Part C: Emerging Technologies, 152, 104191.

Gao H, Liu K, Wang J, & Guo F. (2023) Modular Bus Unit Scheduling for an Autonomous Transit System under Range and Charging Constraints. Applied Sciences. 13, 7661.

Gecchelin, T., & Spera, E. (2022). Selectively combinable independent driving vehicles. U.S. Patent No. 11,535,314. 27 Dec. 2022.

Gecchelin, T., & Webb, J. (2019). Modular dynamic ride-sharing transport systems. Economic Analysis and Policy, 61, 111-117.

Gong, M., Hu, Y., Chen, Z., & Li, X. (2021). Transfer-based customized modular bus system design with passenger-route assignment optimization. Transportation Research Part E: Logistics and Transportation Review, 153, 102422.

Guo, R., Bhatnagar, S., Guan, W., Vallati, M., & Azadeh, S. S. (2023a). Operationalizing modular autonomous customised buses based on different demand prediction scenarios. Transportmetrica A: Transport Science, 2296498.

Guo, R., Guan, W., Vallati, M., & Zhang, W. (2023b). Modular autonomous electric vehicle scheduling for customized on-demand bus services. IEEE Transactions on Intelligent Transportation Systems.

Han, Y., Ma, X., Yu, B., Li, Q., Zhang, R., & Li, X. (2024). Planning two-dimensional trajectories for modular bus enroute docking. Transportation Research Part E: Logistics and Transportation Review, 192, 103769.

Hannoun, G. J., & Menendez, M. (2022). Modular vehicle technology for emergency medical services. Transportation research part C: emerging technologies, 140, 103694.

Hatzenbühler, J., Jenelius, E., Gidófalvi, G., & Cats, O. (2023). Modular vehicle routing for combined passenger and freight transport. Transportation Research Part A: Policy and Practice, 173, 103688.

Huang, D., Hu, Z., Tian, J., & Tu, R. (2025). Improving conventional transit services with modular autonomous vehicles: A bi-level programming approach. Travel Behaviour and Society, 39, 100939.

Ji, Y., Liu, B., Shen, Y., & Du, Y. (2021). Scheduling strategy for transit routes with modular autonomous vehicles. International Journal of Transportation Science and Technology, 10(2), 121-135.

Khan, Z. S., & Menéndez, M. (2023). Bus splitting and bus holding: A new strategy using autonomous modular buses for preventing bus bunching. Transportation Research Part A: Policy and Practice, 177, 103825.

Khan, Z. S., & Menéndez, M. (2024). A seamless bus network without external transfers using autonomous modular vehicles. Transportation Research Part C: Emerging Technologies, 104822.

Khan, Z. S., & Menéndez, M. (2025). No time for stopping: A Stop-Less Autonomous Modular (SLAM) bus service. Transportation Research Part C: Emerging Technologies, 171, 104888.

Khan, Z. S., He, W., & Menéndez, M. (2023). Application of modular vehicle technology to mitigate bus bunching. Transportation Research Part C: Emerging Technologies, 146, 103953.

Li, C., Zhang, M., Jiang, G., & Wang, T. (2025). The bi-mode problem with modular buses and private vehicles in the autonomous driving environment. Transportmetrica B: Transport Dynamics, 13(1), 2440806.

Li, Q., & Li, X. (2022). Trajectory planning for autonomous modular vehicle docking and autonomous vehicle platooning operations. Transportation Research Part E: Logistics and Transportation Review, 166, 102886.

Li, Q., & Li, X. (2023). Trajectory optimization for autonomous modular vehicle or platooned autonomous vehicle split operations. Transportation Research Part E: Logistics and Transportation Review, 176, 103115.

Li, X., & Epureanu, B. I. (2020). AI-based competition of autonomous vehicle fleets with application to fleet modularity. European journal of operational research, 287(3), 856-874.

Lin, J., & Zhang, F. (2024). Modular vehicle-based transit system for passenger and freight co-modal transportation. Transportation Research Part C: Emerging Technologies, 160, 104545.

Lin, J., Nie, Y. M., & Kawamura, K. (2022). An autonomous modular mobility paradigm. IEEE Intelligent Transportation Systems Magazine, 15(1), 378-386.

Lin, X., Chen, Z., Li, M., & He, F. (2024). Bunching-Proof Capabilities of Modular Buses: An Analytical Assessment. Transportation Science.

Liu, T., Ceder, A., & Rau, A. (2020). Using deficit function to determine the minimum fleet size of an autonomous modular public transit system. Transportation Research Record, 2674(11), 532-541.

Liu, X., Qu, X., & Ma, X. (2021). Improving flex-route transit services with modular autonomous vehicles. Transportation Research Part E: Logistics and Transportation Review, 149, 102331.

Liu, Y., Chen, Z., & Wang, X. (2024). Alleviating bus bunching via modular vehicles. Transportation Research Part B: Methodological, 189, 103051.

Liu, Z., de Almeida Correia, G. H., Ma, Z., Li, S., & Ma, X. (2023). Integrated optimization of timetable, bus formation, and vehicle scheduling in autonomous modular public transport systems. Transportation Research Part C: Emerging Technologies, 155, 104306.

Luo, X., Fan, W., Xu, M., & Yan, X. Optimal Design of On-Demand Modular Feeder Transit Services. Available at SSRN 5034582.

Next Future Transportation Incorporation. http://www.next-future-mobility.com/. Accessed September 4, 2024.

Oargă, I. T., Prunean, G., Varga, B. O., Moldovanu, D., & Micu, D. D. (2024b). Comparative Analysis of Energy Efficiency between Battery Electric Buses and Modular Autonomous Vehicles. Applied Sciences, 14(11), 4389.

Oargă, I. T., Varga, B. O., Moldovanu, D., Cărăușan, H., & Prunean, G. (2024a). Modular Autonomous Vehicles' Application in Public Transport Networks: Conceptual Analysis on Airport Connection. Sustainability, 16(4), 1512.

Ohmio. https://ohmio.com/. Accessed September 4, 2024.

Pei, M., Lin, P., Du, J., Li, X., & Chen, Z. (2021). Vehicle dispatching in modular transit networks: A mixed-integer nonlinear programming model. Transportation Research Part E: Logistics and Transportation Review, 147, 102240.

Pei, M., Xu, M., Zhong, L., & Qu, X. (2023). Robust design for underground metro systems with modular vehicles. Tunnelling and Underground Space Technology, 132, 104865.

Rejali, S., Aghabayk, K., Mohammadi, A., & Shiwakoti, N. (2024). Evaluating public a priori acceptance of autonomous modular transit using an extended unified theory of acceptance and use of technology model. Journal of Public Transportation, 26, 100081.

Rezgui, D., Siala, J. C., Aggoune-Mtalaa, W., & Bouziri, H. (2019). Application of a variable

neighborhood search algorithm to a fleet size and mix vehicle routing problem with electric modular vehicles. Computers & Industrial Engineering, 130, 537-550.

Romea, G., & Estrada, M. (2021). Analysis of an autonomous driving modular bus system. Transportation research procedia, 58, 181-188.

Shafiee, H. R. Moghaddam and J. Lin, "Using Autonomous Modular Vehicle Technology as an Alternative for Last-Mile Delivery," 2024 Forum for Innovative Sustainable Transportation Systems (FISTS), Riverside, CA, USA, 2024, pp. 1-6, doi: 10.1109/FISTS60717.2024.10485532.

Shi, X., & Li, X. (2021). Operations design of modular vehicles on an oversaturated corridor with first-in, first-out passenger queueing. Transportation Science, 55(5), 1187-1205.

Shi, X., Chen, Z., Li, X., & Qu, X. (2024). Modular vehicles can reduce greenhouse gas emissions for departure flight baggage transportation. Journal of Air Transport Management, 119, 102633.

Shi, X., Chen, Z., Pei, M., & Li, X. (2020). Variable-capacity operations with modular transits for shared-use corridors. Transportation Research Record, 2674(9), 230-244.

Tang, C., Liu, J., Ceder, A., & Jiang, Y. (2024). Optimisation of a new hybrid transit service with modular autonomous vehicles. Transportmetrica A: transport science, 20(2), 2165424.

The first prototype of the futuristic U-Shift vehicle concept makes its debut https://www.dlr.de/en/latest/news/2020/03/20200917_debut-first-prototype-of-the-futuristic-u-shift-vehicle-concept. Accessed September 4, 2024.

Tian, Q., Lin, Y. H., & Wang, D. Z. (2023). Joint scheduling and formation design for modular-vehicle transit service with time-dependent demand. Transportation Research Part C: Emerging Technologies, 147, 103986.

Tian, Q., Lin, Y. H., Wang, D. Z., & Liu, Y. (2022). Planning for modular-vehicle transit service system: Model formulation and solution methods. Transportation Research Part C: Emerging Technologies, 138, 103627.

Tian, Q., Lin, Y. H., Wang, D. Z., & Yang, K. (2025). Toward real-time operations of modular-vehicle transit services: From rolling horizon control to learning-based approach. Transportation Research Part C: Emerging Technologies, 170, 104938.

Ulrich, C., Friedrich, H. E., Weimer, J., & Schmid, S. A. (2019). New operating strategies for an on-the-road modular, electric and autonomous vehicle concept in urban transportation. World Electric Vehicle Journal, 10(4), 91.

Wang, T., Jian, S., Zhou, C., Jia, B., & Long, J. (2024). Multimodal traffic assignment considering heterogeneous demand and modular operation of shared autonomous vehicles. Transportation Research Part C: Emerging Technologies, 169, 104881.

Wang, Y., Ceder, A., Cao, Z., & Zhang, S. (2023). Optimal public transport timetabling with autonomous-vehicle units using coupling and decoupling tactics. Transportmetrica A: Transport Science, 1-51.

What is the Mercedes-Benz Vision URBANETIC? https://www.mbscottsdale.com/blog/what-is-the-mercedes-benz-vision-urbanetic/. Accessed September 4, 2024.

Wu, J., Kulcsár, B., & Qu, X. (2021). A modular, adaptive, and autonomous transit system (MAATS): An in-motion transfer strategy and performance evaluation in urban grid transit networks. Transportation Research Part A: Policy and Practice, 151, 81-98.

Xia, D., Ma, J., & Azadeh, S. S. (2024). Integrated timetabling, vehicle scheduling, and dynamic capacity allocation of modular autonomous vehicles under demand uncertainty. arXiv preprint arXiv:2410.16409.

Xia, D., Ma, J., Azadeh, S. S., & Zhang, W. (2023). Data-driven distributionally robust timetabling and dynamic-capacity allocation for automated bus systems with modular vehicles. Transportation Research Part C: Emerging Technologies, 155, 104314.

Ye, L., & Yamamoto, T. (2024). Modular Autonomous Vehicle in Heterogeneous Traffic Flow: Modeling, Simulation, and Implication. arXiv preprint arXiv:2409.17945.

Yi, H., Liu, Y., Menendez, M., & Tang, L. Charging Scheduling and Route Planning for Modular Bus Systems Considering Non-Linear Charging Profile. Available at SSRN 4785233.

Yi, H., Liu, Y., Yang, H., & Menéndez, M. (2024). Charging-on-the-move Public Transit Systems Constituted by Modular Trolleybuses and Modular Buses. IEEE Transactions on Transportation Electrification.

Zermasli, D., Iliopoulou, C., Laskaris, G., & Kepaptsoglou, K. (2023). Feeder bus network design with modular transit vehicles. Journal of Public Transportation, 25, 100078.

Zhang, J., Ge, Y. E., Tang, C., & Zhong, M. (2024). Optimising modular-autonomous-vehicle transit service employing coupling–decoupling operations plus skip-stop strategy. Transportation Research Part E: Logistics and Transportation Review, 184, 103450.

Zhang, Z., Tafreshian, A., & Masoud, N. (2020). Modular transit: Using autonomy and modularity to improve performance in public transportation. Transportation Research Part E: Logistics and Transportation Review, 141, 102033.

Zhou, H., Li, Y., Ma, C., Long, K., & Li, X. (2024). Modular Vehicle Routing Problem: Applications in Logistics. arXiv preprint arXiv:2409.01518.

Zou, K., Zhang, K., & Li, M. (2024). Operational design for modular electrified transit in corridor areas. Transportation Research Part E: Logistics and Transportation Review, 187, 103567.